\documentclass[10pt,twocolumn,letterpaper]{article}

\usepackage{cvpr}
\usepackage{times}
\usepackage{epsfig}
\usepackage{graphicx}
\usepackage{amsmath}
\usepackage{amssymb}
\usepackage{subfigure}
\usepackage{tabularx}
\usepackage{color}
\usepackage{multirow}
\usepackage{colortbl}

\newcommand{\figref}[1]{Fig. \ref{#1}}
\newcommand{\tabref}[1]{Table \ref{#1}}
\newcommand{\equref}[1]{(\ref{#1})}

\makeatletter
\def\hlinewd#1{%
	\noalign{\ifnum0=`}\fi\hrule \@height #1 \futurelet
	\reserved@a\@xhline}
\usepackage[breaklinks=true,bookmarks=false]{hyperref}

\cvprfinalcopy % *** Uncomment this line for the final submission

\def\cvprPaperID{****} % *** Enter the CVPR Paper ID here

\begin{document}

%%%%%%%%% TITLE
\title{FCSS: Fully Convolutional Self-Similarity for Dense Semantic Correspondence}

\author{Seungryong Kim\textsuperscript{1}, Dongbo Min\textsuperscript{2}, Bumsub Ham\textsuperscript{1}, Sangryul Jeon\textsuperscript{1}, Stephen Lin\textsuperscript{3}, Kwanghoon Sohn\textsuperscript{1}\\
	{\textsuperscript{1}Yonsei University \quad \textsuperscript{2}Chungnam National University \quad \textsuperscript{3}Microsoft Research}\\
	{\tt\small \{srkim89,mimo,cheonjsr,khsohn\}@yonsei.ac.kr \quad dbmin@cnu.ac.kr \quad stevelin@microsoft.com}}

\maketitle
%\thispagestyle{empty}

%%%%%%%%% ABSTRACT
\begin{abstract}
	We present a descriptor, called fully convolutional self-similarity (FCSS), for dense semantic correspondence. To robustly match points among different instances within the same object class, we formulate FCSS using local self-similarity (LSS) within a fully convolutional network. In contrast to existing CNN-based descriptors, FCSS is inherently insensitive to intra-class appearance variations because of its LSS-based structure, while maintaining the precise localization ability of deep neural networks. The sampling patterns of local structure and the self-similarity measure are jointly learned within the proposed network in an end-to-end and multi-scale manner. As training data for semantic correspondence is rather limited, we propose to leverage object candidate priors provided in existing image datasets and also correspondence consistency between object pairs to enable weakly-supervised learning. Experiments demonstrate that FCSS outperforms conventional handcrafted descriptors and CNN-based descriptors on various benchmarks.
\end{abstract}

\section{Introduction}\label{sec:1}
\begin{figure}
	\centering
	\renewcommand{\thesubfigure}{}
	\subfigure[(a) Source image]
	{\includegraphics[width=0.498\linewidth]{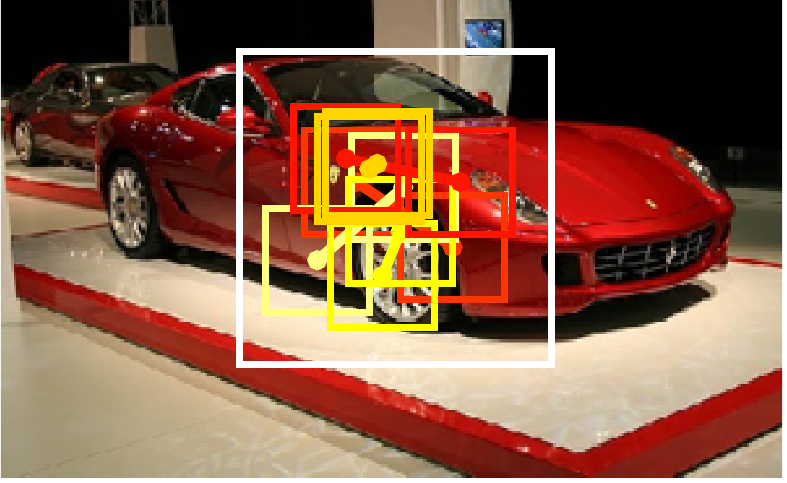}}\hfill
	\subfigure[(b) Target image]
	{\includegraphics[width=0.498\linewidth]{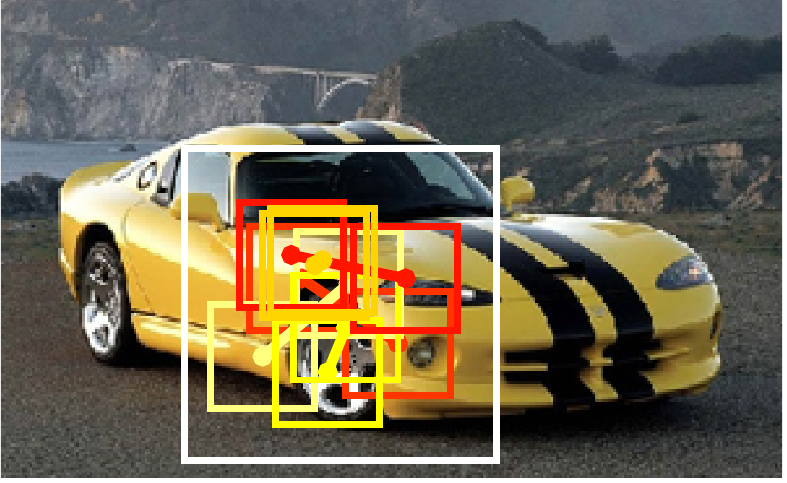}}\hfill
	\vspace{-6pt}
	\subfigure[(c) Window]
	{\includegraphics[width=0.245\linewidth]{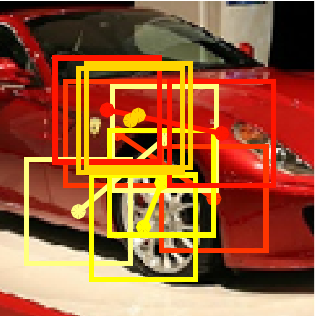}}\hfill
	\subfigure[(d) Window]
	{\includegraphics[width=0.245\linewidth]{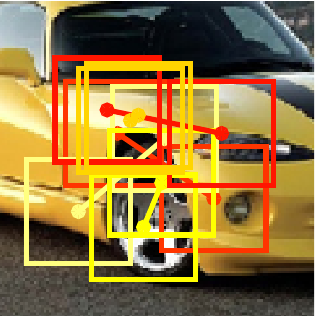}}\hfill
	\subfigure[(e) FCSS in (c)]
	{\includegraphics[width=0.245\linewidth]{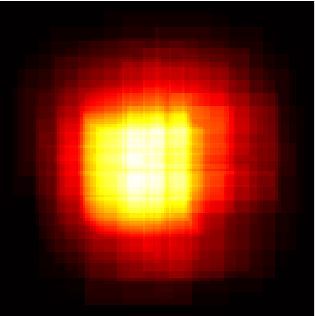}}\hfill
	\subfigure[(f) FCSS in (d)]
	{\includegraphics[width=0.245\linewidth]{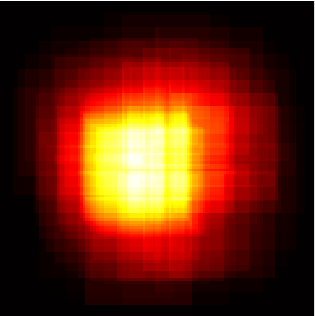}}\hfill
	\vspace{-1pt}
	\caption{Visualization of local self-similarity. Even though there are significant differences in appearance among different instances within the same object class in (a) and (b), local self-similarity in our FCSS descriptor is preserved between them as shown in (e) and (f), thus providing robustness to intra-class variations.}\label{img:1}\vspace{-10pt}
\end{figure}

Establishing dense correspondences across
\emph{semantically} similar images is essential for numerous tasks
such as scene recognition, image registration, semantic
segmentation, and image editing
\cite{HaCohen2011,Liu11,Kim13,Yang14,Zhou15}. Unlike traditional
dense correspondence approaches for estimating depth
\cite{Scharstein02} or optical flow \cite{Butler12,Sun10}, in which
\emph{visually} similar images of the same scene are used
as inputs, semantic correspondence estimation poses additional challenges
due to intra-class variations among object instances,
as exemplified in \figref{img:1}.

Often, basic visual properties such as colors and gradients
are not shared among different object instances
in the same class. These variations, in addition to other complications
from occlusion and background clutter, lead to significant differences
in appearance that can distract matching by
handcrafted feature descriptors \cite{Lowe04,Tola10}. Although
powerful optimization techniques can help by enforcing smoothness constraints over a correspondence map \cite{Liu11,Kim13,Zhou15,Taniai16,Ham16}, they are limited in effectiveness
without a proper matching descriptor for semantic correspondence estimation.

Over the past few years, convolutional neural network (CNN) based features
have become increasingly popular for correspondence estimation thanks to their localization precision of matched points and their invariance to minor geometric deformations and illumination changes \cite{Han15,Zagoruyko15,Simo-Serra15b,Yi16}.
However, for computing semantic correspondences within this framework,
greater invariance is needed to deal with the more substantial
appearance differences. This could potentially be achieved with a deeper convolutional network \cite{Simonyan15}, but would come at the cost of significantly reduced
localization precision in matching details as shown in \cite{Long15,Hariharan15}.
Furthermore, as training data for semantic
correspondence is rather limited, a network cannot be trained properly
in a supervised manner.

To address these issues, we introduce a CNN-based descriptor that is
inherently insensitive to intra-class appearance variations while
maintaining precise localization ability. The key insight, illustrated
in \figref{img:1}, is that among different object instances in the same class,
their local structural layouts remain roughly the same. Even with dissimilar
colors, gradients, and small differences in feature positions, the local
self-similarity (LSS) between sampled patch pairs is basically preserved.
This property has been utilized for non-rigid object detection
\cite{Schechtman07}, sketch retrieval \cite{Chatfield09},
and cross-modal correspondence \cite{Kim15}. However, existing
LSS-based techniques are mainly handcrafted and need further robustness to capture reliable matching evidence from
semantically similar images.

Our proposed descriptor, called fully convolutional
self-similarity (FCSS), formulates LSS within a fully convolutional
network in manner where the patch sampling patterns and self-similarity
measure are both learned. We propose a
convolutional self-similarity (CSS) layer that encodes the LSS structure
and possesses differentiability, allowing for end-to-end training
together with the sampling patterns. The convolutional self-similarities
are measured at multiple scales, using skip layers \cite{Long15} to forward intermediate
convolutional activations. Furthermore, since limited training data is
available for semantic correspondence, we propose a weakly-supervised feature learning
scheme that leverages correspondence consistency between object locations
provided in existing image datasets. Experimental results show that the FCSS descriptor
outperforms conventional handcrafted descriptors and CNN-based
descriptors on various benchmarks, including that of Taniai
\emph{et al.}~\cite{Taniai16}, Proposal Flow~\cite{Ham16}, the PASCAL dataset~\cite{Chen14}, and Caltech-101~\cite{Fei-Fei06}.

\section{Related Work}\label{sec:2}
\paragraph{Feature Descriptors}\label{sec:21}
Conventional gradient-based and intensity
comparison-based descriptors, such as SIFT \cite{Lowe04}, HOG \cite{Dalal05}, DAISY \cite{Tola10}, and BRIEF \cite{Calonder11},
have shown limited performance in dense correspondence estimation
across semantically similar but different object instances. Over
the past few years, besides these handcrafted features, several
attempts have been made using deep CNNs to learn discriminative
descriptors for local patches from large-scale datasets. Some of
these techniques have extracted immediate activations as the
descriptor \cite{Gong14,Fischer14,Donahue14,Long14}, which have
shown to be effective for patch-level matching. Other methods have
directly learned a similarity measure for comparing patches using a
convolutional similarity network \cite{Han15, Zagoruyko15, Simo-Serra15b, Yi16}. Even though CNN-based descriptors encode a
discriminative structure with a deep architecture, they have
inherent limitations in handling large intra-class variations \cite{Simo-Serra15b,Dong15}.
Furthermore, they are mostly tailored to estimate sparse
correspondences, and cannot in practice provide dense descriptors
due to their high computational complexity. Of particular
importance, current research on semantic correspondence lacks an
appropriate benchmark with dense ground-truth correspondences, making
supervised learning of CNNs less feasible for this task.

LSS techniques, originally proposed in \cite{Schechtman07}, have achieved impressive results in object detection, image retrieval by
sketching \cite{Schechtman07}, deformable shape class retrieval
\cite{Chatfield09}, and cross-modal correspondence estimation 
\cite{Torabi13,Kim15}. Among the more recent cross-modal descriptors is the dense
adaptive self-correlation (DASC) descriptor~\cite{Kim15}, which
provides satisfactory performance but is unable to handle non-rigid
deformations due to its fixed patch pooling scheme. The deep
self-correlation (DSC) descriptor~\cite{Kim16} reformulates LSS in a
deep non-CNN architecture. As all of these techniques utilize handcrafted descriptors, they lack the robustness that is possible with CNNs. \vspace{-10pt}

\paragraph{Dense Semantic Correspondence}\label{sec:22}
Many techniques for dense semantic correspondence employ handcrafted features such as SIFT~\cite{Lowe04} or HOG~\cite{Dalal05}. To improve the matching quality, they focus on optimization. Among these methods are some based on SIFT Flow \cite{Liu11,Kim13}, which uses hierarchical dual-layer belief propagation (BP). Other instances include the methods with an exemplar-LDA approach \cite{Bristow15}, through joint image set alignment \cite{Zhou15}, or together with cosegmentation~\cite{Taniai16}.

More recently, more powerful CNN-based descriptors have been used for establishing dense semantic correspondences. Pre-trained ConvNet features~\cite{Krizhevsky12} were employed with the SIFT Flow algorithm~\cite{Long14} and with semantic flow using object proposals~\cite{Ham16}. Choy \emph{et al.} \cite{Choy16} proposed a
deep convolutional descriptor based on fully convolutional feature
learning and a convolutional spatial transformer \cite{Jaderberg15}.    As these methods formulate the networks by combining existing convolutional
networks only, they face a tradeoff between appearance
invariance and localization precision that presents inherent limitations on semantic correspondence. \vspace{-10pt}

\paragraph{Weakly-Supervised Feature Learning}\label{sec:23}
For the purpose of object recognition, Dosovitskiy \emph{et al.} \cite{Dosovitskiy14} trained the network to discriminate between a set of surrogate classes formed by applying various transformations.
For object matching, Lin \emph{et al.} \cite{Lin16} proposed an
unsupervised learning to learn a compact binary
descriptor by leveraging an iterative training scheme. More closely
related to our work is the method of Zhou \emph{et al.} \cite{Zhou16}, which exploits cycle-consistency with a 3D CAD model
\cite{ShapeNet} as a supervisory signal to train a deep network for
semantic correspondence. However, the need to have a suitable 3D CAD
model for each object class limits its applicability.

\section{The FCSS Descriptor}\label{sec:3}
\subsection{Problem Formulation and Overview}\label{sec:31}
Let us define an image $I$ such that $I_{i}:\mathcal{I} \to {\mathbb R}^3$ for pixel $i=[i_\mathbf{x},i_\mathbf{y}]^{T}$.
For each image point $I_{i}$, a dense descriptor ${D_i}:\mathcal{I} \to \mathbb{R}^L$ of dimension $L$ is defined on a local support window.
For LSS, this descriptor represents locally self-similar structure around a given pixel
by recording the similarity between certain patch pairs within a local window.
Formally, LSS can be described as a vector of feature values
$D_{i} = { \bigcup_{l}}D_{i} (l)$ for $l \in \{1,...,L\}$, where the feature values are computed as
\begin{equation}\label{equ:self-similar}
D_{i} (l) = \mathop {\mathrm{max}}\nolimits_{j \in \mathcal{N}_i}
\mathrm{exp} \left(-\mathcal{S}\left(P_{j-s_{l}},P_{j-t_{l}}\right) /
\lambda \right),
\end{equation}
where $\mathcal{S}(P_{i-s_{l}},P_{i-t_{l}})$ is a self-similarity distance between two patches $P_{i-s_{l}}$ and $P_{i-t_{l}}$ sampled on $s_{l}$ and $t_{l}$, the $l^{th}$ selected sampling pattern, around center pixel $i$. To alleviate the effects of outliers,
the self-similarity responses are encoded by non-linear mapping with an exponential function of a bandwidth $\lambda$ \cite{Black98}. For spatial invariance to the position of the sampling pattern,
the maximum self-similarity within a spatial window $\mathcal{N}_i$ is computed.

By leveraging CNNs, our objective is to design a dense descriptor
that formulates LSS in a fully convolutional and end-to-end manner for robust estimation of dense semantic correspondences.
Our network is built as a multi-scale series of convolutional self-similarity (CSS) layers that each includes a two-stream shifting transformer for applying a sampling pattern. To learn the network, including its self-similarity measures and sampling patterns, in a weakly-supervised manner, our network utilizes correspondence consistency between pairs of input images as well as object locations provided in existing datasets.

\subsection{CSS: Convolutional Self-Similarity Layer}\label{sec:32}
We first describe the convolutional self-similarity (CSS) layer, which provides robustness to intra-class variations while preserving localization precision of matched points around fine-grained object boundaries.\vspace{-10pt}

\begin{figure}
	\centering
	\renewcommand{\thesubfigure}{}
	\subfigure[(a) Straightforward implementation of CSS layer]
	{\includegraphics[width=1\linewidth]{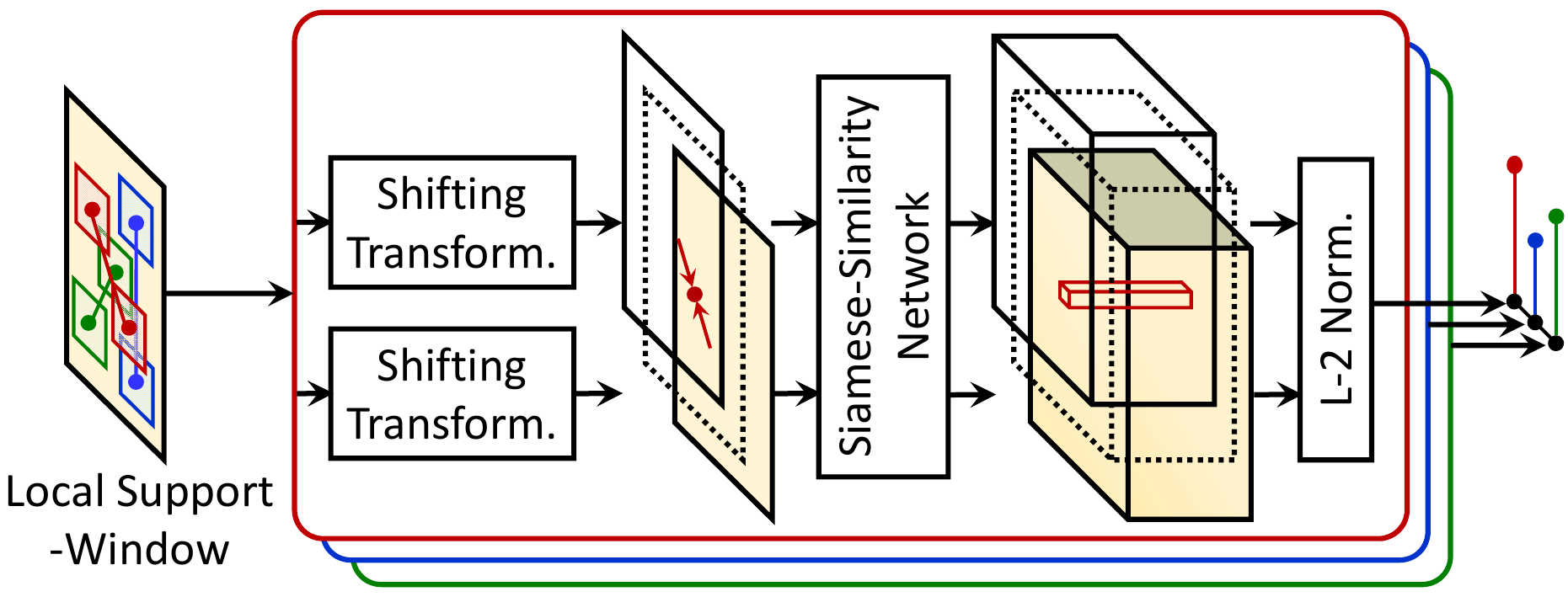}}\\
	\vspace{-6pt}
	\subfigure[(b) Efficient implementation of CSS layer]
	{\includegraphics[width=1\linewidth]{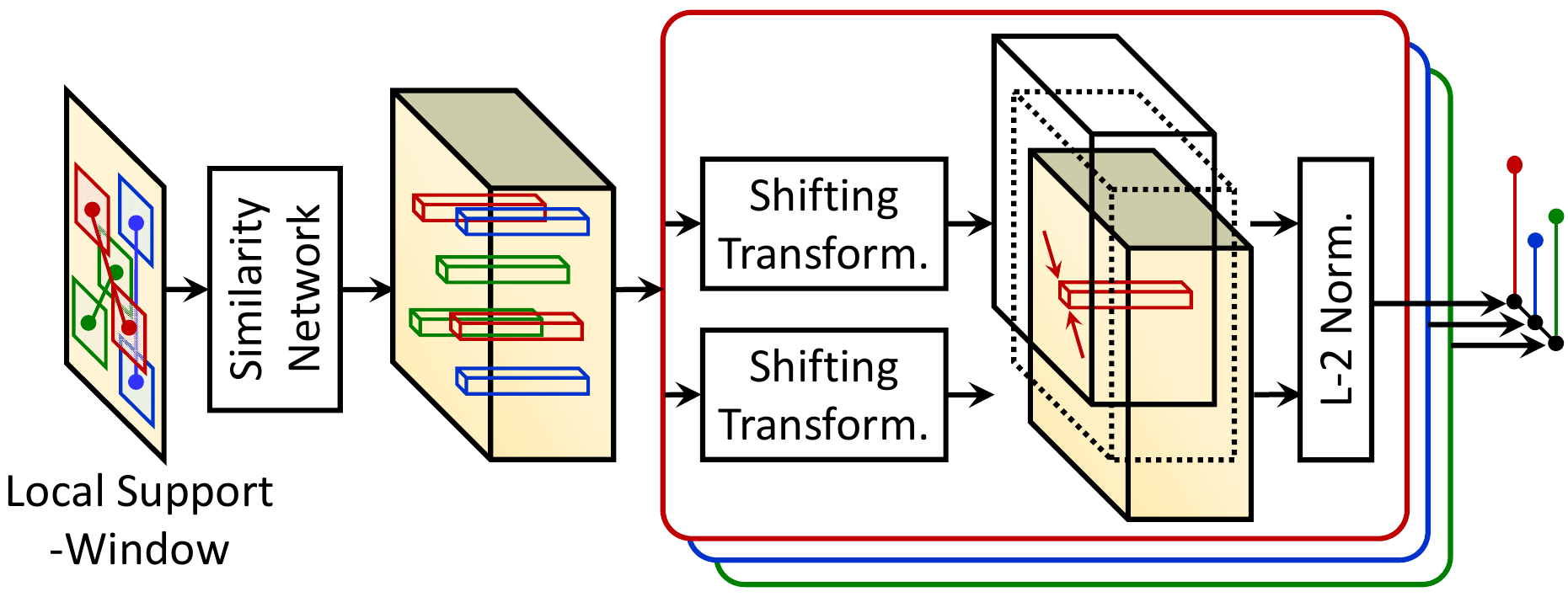}}\\
	\vspace{-1pt}
	\caption{Convolutional self-similarity (CSS) layers, implemented as
		(a) straightforward and (b) efficient versions. With the efficient scheme, convolutional self-similarity is equivalently solved while avoiding repeated computations for convolutions.
	}\label{img:2}\vspace{-10pt}
\end{figure}

\paragraph{Convolutional Similarity Network}
Previous LSS-based techniques \cite{Schechtman07,Kim15,Kim16} evaluate \equref{equ:self-similar} by
sampling patch pairs and then computing their similarity using
handcrafted metrics, which often fails to yield detailed matching
evidence for estimating semantic correspondences. Instead, we compute
the similarity of sampled patch pairs through CNNs.
With $l$ omitted for simplicity, the self-similarity between a
patch pair $P_{i-s}$ and $P_{i-t}$ is formulated through a Siamese
network, followed by decision or metric network \cite{Zagoruyko15,Han15} or
a simple $L_2$ distance \cite{Simo-Serra15b,Yi16} as shown in
\figref{img:2}(a). Specifically, convolutional activations through
feed-forward processes $\mathcal{F} (P_{i-s};\mathbf{W}_c)$ and
$\mathcal{F} (P_{i-t};\mathbf{W}_c)$ with CNN parameters
$\mathbf{W}_c$ are used to measure self-similarity based on the $L_2$
distance, such that
\begin{equation}\label{equ:css-brute}
\mathcal{S}(P_{i-s},P_{i-t}) =
\|\mathcal{F} (P_{i-s};\mathbf{W}_c) - \mathcal{F} (P_{i-t};\mathbf{W}_c)\|^2.
\end{equation}

Note that our approach employs the Siamese network to measure \emph{self-similarity within a single image}, 
in contrast to recent CNN-based descriptors \cite{Simo-Serra15b} that directly
measure the \emph{similarity between patches from two different images}.

However, computing $\mathcal{S}(P_{i-s},P_{i-t})$ for all sampling patterns $(s,t)$
in this network is time-consuming, since the number of iterations through
the Siamese network is linearly proportional to the number of sampling
patterns. To expedite this computation, we instead generate the
convolutional activations of an entire image by passing it through the CNN, 
similar to \cite{He15}, and then measure the self-similarity for the sampling patterns directly on the
convolutional activations $\mathbf{A}_i = \mathcal{F}
(I_i;\mathbf{W}_c)$, as shown in \figref{img:2}(b). Formally, this can
be written as
\begin{equation}\label{equ:css-efficient}
\mathcal{S}(P_{i-s},P_{i-t}) = \|\mathbf{A}_{i-s} -
\mathbf{A}_{i-t}\|^2.
\end{equation}

With this scheme, the self-similarity is measured by running the
similarity network only once, regardless of the number of sampling patterns.
Interestingly, a similar computational scheme was used to measure
the similarity between two different images in \cite{Zbontar16},
whereas our scheme instead measures self-similarity within a single
image. \vspace{-10pt}

\paragraph{Two-Stream Shifting Transformer}
The sampling patterns $(s,t)$ of patch pairs are a critical element of local
self-similarity. In our CSS layer, a sampling pattern for a pixel $i$
can be generated by shifting the original activation $\mathbf{A}_i$
by $s$ and $t$ to form two different activations from which
self-similarity is measured. While this spatial manipulation of
data within the network could be learned and applied using a
spatial transformer layer~\cite{Jaderberg15}, we instead formulate a
simplification of this, called a shifting transformer layer, in which
the shift transformations $s$ and $t$ are defined as network parameters
that can be learned because of the differentiability of the shifting
transformer layer. In this way, the optimized sampling patterns can be
learned in the CNN.

Concretely, the sampling patterns are defined as network parameters
$\mathbf{W}_s=[\mathbf{W}_{s_\mathbf{x}},\mathbf{W}_{s_\mathbf{y}}]^{T}$
and
$\mathbf{W}_t=[\mathbf{W}_{t_\mathbf{x}},\mathbf{W}_{t_\mathbf{y}}]^{T}$
for all $(s,t)$.
Since the shifted sampling is repeated in an (integer) image
domain, the convolutional self-similarity activation $\mathbf{A}_i$ is
shifted simply without interpolation in the image domain according to the sampling patterns.
We first define the sampled activations though a two-stream
shifting transformer as
\begin{equation}\label{equ:css-sft}
\mathbf{A}_{i-\mathbf{W}_s} = \mathcal{F} (\mathbf{A}_{i};\mathbf{W}_s),\quad
\mathbf{A}_{i-\mathbf{W}_t} = \mathcal{F} (\mathbf{A}_{i};\mathbf{W}_t).
\end{equation}
From this, convolutional self-similarity is then defined as
\begin{equation}\label{equ:css-efficient-sft}
\mathcal{S}(P_{i-\mathbf{W}_s},P_{i-\mathbf{W}_t}) = \|\mathcal{F}
(\mathbf{A}_i;\mathbf{W}_s) - \mathcal{F}
(\mathbf{A}_i;\mathbf{W}_t)\|^2.
\end{equation}
Note that $\mathcal{S}(P_{i-\mathbf{W}_s},P_{i-\mathbf{W}_t})$
represents a convolutional self-similarity vector defined
for all $(s,t)$. \vspace{-10pt}

\paragraph{Differentiability of Convolutional Self-Similarity}
For end-to-end learning of the proposed descriptor, the derivatives for
the CSS layer must be computable, so that
gradients of the final loss can be back-propagated to the
convolutional similarity and
shifting transformer layers.

To obtain the derivatives for the convolutional similarity layer and
the shifting transformer layers, we first compute the
Taylor expansion of the shifting transformer activations,
under the assumption that $\mathbf{A}_i$ is smoothly varying with
respect to shifting parameters $\mathbf{W}_s$:
\begin{equation}\label{equ:sft-taylor}
\mathbf{A}_{i-\mathbf{W}_s^n} = \mathbf{A}_{i-\mathbf{W}_s^{n-1}}
+ (\mathbf{W}_s^n-\mathbf{W}_s^{n-1}) \circ \triangledown \mathbf{A}_{i-\mathbf{W}_s^{n-1}},
\end{equation}
where $\mathbf{W}_s^{n-1}$ represents the sampling patterns at the
$(n-1)^{th}$ iteration during training, and $\circ$ denotes the
Hadamard product. $\triangledown \mathbf{A}_{i-\mathbf{W}_s^{n-1}}$
is a spatial derivative on each activation slice with respect to
$\triangledown_\mathbf{x}$ and $\triangledown_\mathbf{y}$. By
differentiating \equref{equ:sft-taylor} with respect to
$\mathbf{W}^n_{s_\mathbf{x}}$, we get the shifting parameter
derivatives as
\begin{equation}\label{equ:deriv-sft}
\frac{\partial \mathbf{A}_{i-\mathbf{W}_s^n}}{\partial \mathbf{W}^n_{s_\mathbf{x}}}
= \triangledown_\mathbf{x} \mathbf{A}_{i-\mathbf{W}_s^{n-1}}.
\end{equation}

By the chain rule, with $n$ omitted, the derivative of the final loss
$\mathcal{L}$ with respect to $\mathbf{W}_{s_\mathbf{x}}$ can be expressed as
\begin{equation}\label{equ:deriv-param}
\frac{\partial \mathcal{L}}{\partial \mathbf{W}_{s_\mathbf{x}}}
= \frac{\partial \mathcal{L}}{\partial \mathbf{A}_{i-\mathbf{W}_s}}
\frac{\partial \mathbf{A}_{i-\mathbf{W}_s}}{\partial \mathbf{W}_{s_\mathbf{x}}}.
\end{equation}
Similarly, $\partial \mathcal{L}/\partial
\mathbf{W}_{s_\mathbf{y}}$, $\partial \mathcal{L}/\partial
\mathbf{W}_{t_\mathbf{x}}$, and $\partial \mathcal{L}/\partial
\mathbf{W}_{t_\mathbf{y}}$ can be calculated.

Moreover, the derivative of the final loss $\mathcal{L}$ with respect
to $\mathbf{A}_{i}$ can be formulated as
\begin{equation}\label{equ:deriv-input}
\begin{split}
\frac{\partial \mathcal{L}}{\partial \mathbf{A}_{i}}
&= \frac{\partial\mathcal{L}}{\partial \mathbf{A}_{i-\mathbf{W}_s}}
\frac{\partial \mathbf{A}_{i-\mathbf{W}_s}}{\partial \mathbf{A}_{i}}
+ \frac{\partial\mathcal{L}}{\partial \mathbf{A}_{i-\mathbf{W}_t}}
\frac{\partial \mathbf{A}_{i-\mathbf{W}_t}}{\partial \mathbf{A}_{i}}\\
&= \frac{\partial\mathcal{L}}{\partial \mathbf{A}_{i-\mathbf{W}_s}}
+ \frac{\partial\mathcal{L}}{\partial \mathbf{A}_{i-\mathbf{W}_t}},
\end{split}
\end{equation}
since $\partial \mathbf{A}_{i-\mathbf{W}_s}/\partial \mathbf{A}_{i}$ is 1 on the pixel ${i-\mathbf{W}_s}$. In this way, the derivatives for the CSS layer
can be computed.
\begin{figure}
	\centering
	\renewcommand{\thesubfigure}{}
	\subfigure[]
	{\includegraphics[width=1\linewidth]{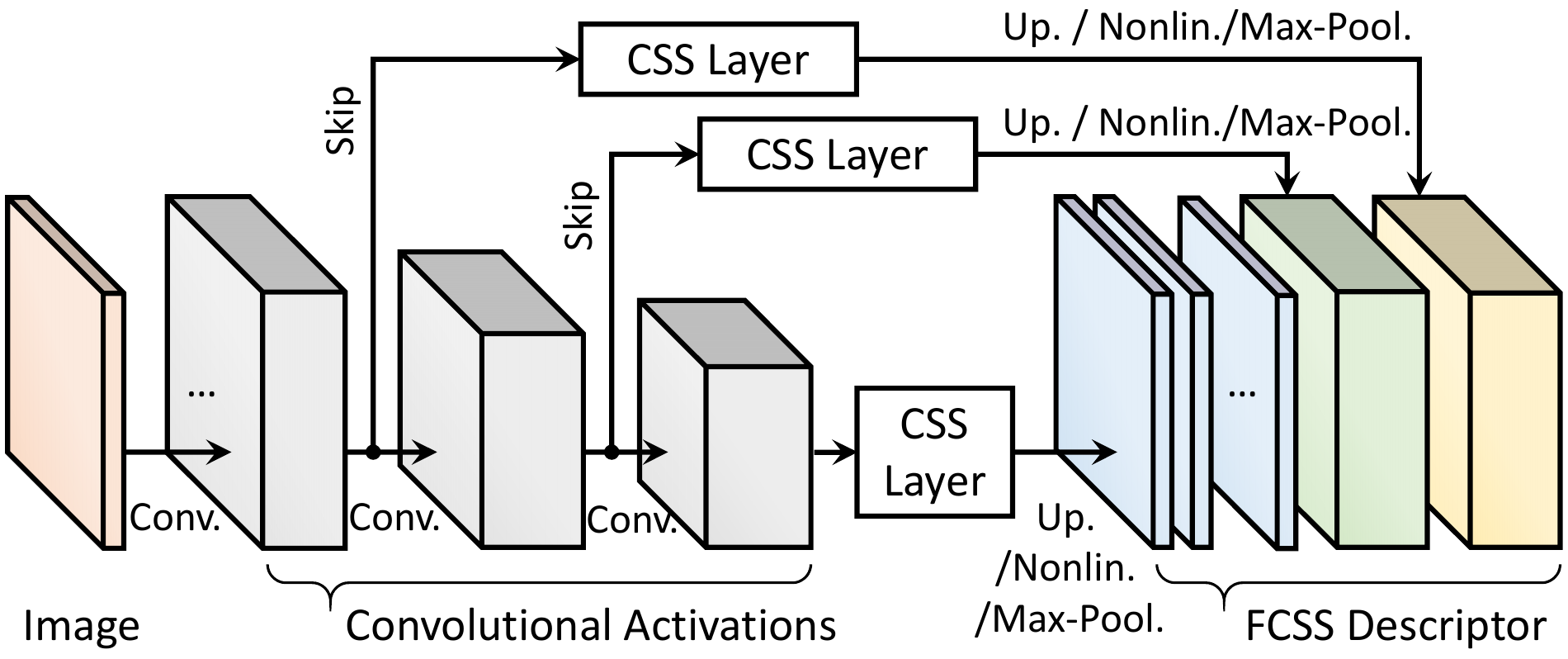}}\\
	\vspace{-12pt}
	\caption{Network configuration of the FCSS descriptor, consisting of convolutional self-similarity layers at multiple scales.}\label{img:3}\vspace{-10pt}
\end{figure}

\subsection{Network Configuration for Dense Descriptor}\label{sec:33}
\paragraph{Multi-Scale Convolutional Self-Similarity Layer}
In building the descriptor through a CNN architecture, there is a
trade-off between robustness to semantic variations
and fine-grained localization precision \cite{Long15,Hariharan15}. The deeper convolutional layers gain greater robustness to semantic variations, but also lose localization precision of matching details around object
boundaries. On the contrary, the shallower convolutional layers
better preserve matching details, but are more sensitive to intra-class
appearance variations.

Inspired by the skip layer scheme in \cite{Long15}, we formulate the
CSS layers in a hierarchical manner to encode multi-scale
self-similarities as shown in \figref{img:3}. Even though the CSS layer itself provides robustness to semantic variations and fine-grained localization precision, this scheme enables
the descriptor to boost both robustness and localization precision. The CSS layers are located after multiple intermediate activations, and
their outputs are concatenated to construct the proposed descriptor.
In this way, the descriptor naturally encodes self-similarity at
multiple scales of receptive fields, and further learns optimized sampling
patterns on each scale. Many existing descriptors \cite{Hariharan15,Zagoruyko15} also
employ a multi-scale description to improve matching quality.

For intermediate activations $\mathbf{A}^k_i = \mathcal{F}
(I_i;\mathbf{W}^k_c)$, where $k \in \{1,...,K\}$ is the level of
convolutional activations and $\mathbf{W}^k_c$ is convolutional similarity network parameters at the $k^{th}$ level, the self-similarity at the
the $k^{th}$ level is measured according to sampling patterns $\mathbf{W}^k_s$ and $\mathbf{W}^k_t$ as
\begin{equation}\label{equ:multi-css}
\mathcal{S}(P_{i-\mathbf{W}^k_s},P_{i-\mathbf{W}^k_t}) =
\|\mathcal{F} (\mathbf{A}^k_i;\mathbf{W}^k_s) - \mathcal{F} (\mathbf{A}^k_i;\mathbf{W}^k_t)\|^2.
\end{equation}

Since the intermediate activations are of smaller spatial
resolutions than the original image resolution,
we apply a bilinear upsampling layer \cite{Long15} after each CSS
layer. \vspace{-10pt}

\paragraph{Non-linear Gating and Max-Pooling Layer}
The CSS responses are passed through a non-linear gating layer to
mitigate the effects of outliers \cite{Black98}. Furthermore, since the pre-learned sampling patterns used in the CSS layers are fixed over
an entire image, they may be sensitive to non-rigid deformation as
described in \cite{Kim16}. To address this, we perform the
max-pooling operation within a spatial window $\mathcal{N}_i$ centered at a
pixel $i$ after the non-linear gating:
\begin{equation}\label{equ:ucss-descriptor}
D^{k}_{i} = \mathop {\mathrm{max}}\nolimits_{j\in \mathcal{N}_i} \mathrm{exp} ( - \mathcal{S} (P_{j-\mathbf{W}^k_s},P_{j-\mathbf{W}^k_t})
/\mathbf{W}^k_\lambda ),
\end{equation}
where $\mathbf{W}^k_\lambda$ is a learnable parameter for scale $k$.
The max-pooling layer provides an effect similar to using
pixel-varying sampling patterns, providing robustness to non-rigid deformation. The descriptor for each pixel then undergoes $L_2$
normalization. Finally, the proposed descriptor $D_{i} = {\bigcup_{k}} D^{k}_{i}$ is built by concatenating feature responses
across all scales. \figref{img:3} displays an overview of the FCSS descriptor
construction.

\subsection{Weakly-Supervised Dense Feature Learning}\label{sec:34}
A major challenge of semantic correspondence estimation with CNNs is
the lack of ground-truth correspondence maps for training data. Constructing training data without manual annotation is
difficult due to the need for semantic understanding. Moreover,
manual annotation is very labor intensive and somewhat subjective.
To deal with this problem, we propose a weakly-supervised learning scheme based on correspondence consistency between image
pairs.\vspace{-10pt}

\paragraph{Fully Convolutional Feature Learning}
For training the network with image pairs $I$ and $I'$, the
correspondence contrastive loss \cite{Choy16} is defined as
\begin{equation}\label{equ:final-loss}
\begin{split}
\mathcal{L}(\mathbf{W}) &=
\frac{1}{2N}\sum\nolimits_{i\in\Omega}{l_i \| \mathcal{F} (I_{i};\mathbf{W}) - \mathcal{F} (I'_{i'};\mathbf{W}) \|^2}\\
&{(1-l_i) \mathrm{max}(0,C-\| \mathcal{F} (I_{i};\mathbf{W}) -
	\mathcal{F} (I'_{i'};\mathbf{W}) \|^2)},
\end{split}
\end{equation}
where $i$ and $i'$ are either a matching or non-matching pixel pair, 
and $l_i$ denotes a class label that is 1 for a positive pair and
0 otherwise. 
$\Omega$ represents the set of training samples, and $N$ is the number of training samples. 
$C$ is the maximal cost. The loss for a
negative pair approaches zero as their distance increases.
$\mathbf{W} =
\{\mathbf{W}^k_c,\mathbf{W}^k_s,\mathbf{W}^k_t,\mathbf{W}^k_\lambda
\;|\; k = 1,...,K\}$ represents all network parameters. By
back-propagating the partial derivative of
$\mathcal{L}(\mathbf{W})$, the overall network can be
learned.

Unlike existing CNN-based descriptor learning methods which use a set of \emph{patch pairs}~\cite{Simo-Serra15b,Zagoruyko15,Han15}, we use a set of \emph{image pairs} for training. Such an image-wise learning scheme expedites feature learning by reducing the computational redundancy that occurs when computing convolutional activations for two adjacent pixels in the image. Our approach is conceptually similar to \cite{Choy16}, but we learn the descriptor in a weakly-supervised manner that leverages correspondence consistency between each image pair so that the positive and negative samples are actively determined during training. \vspace{-10pt}
\begin{figure}
	\centering
	\renewcommand{\thesubfigure}{}
	\subfigure[]
	{\includegraphics[width=1\linewidth]{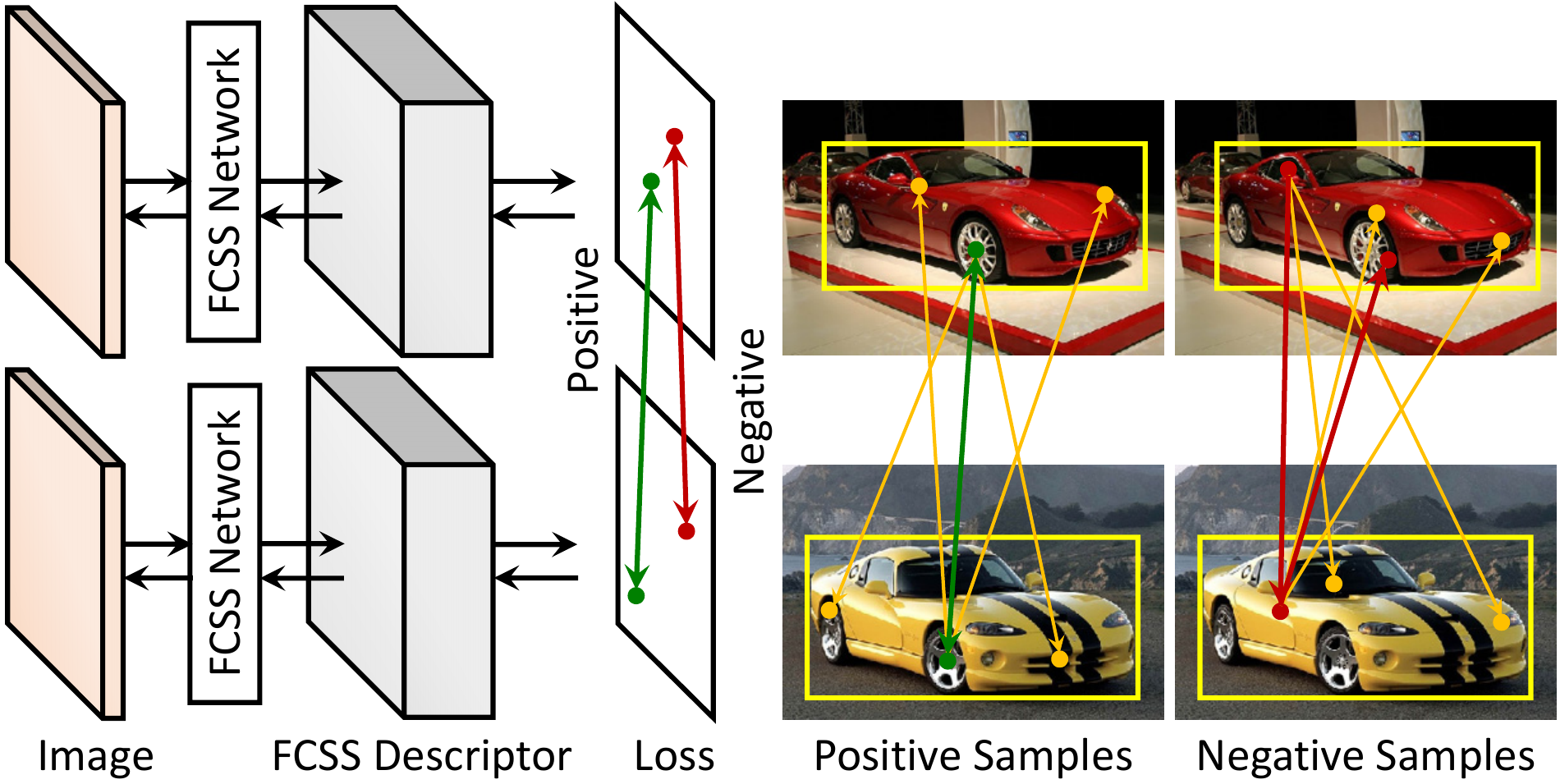}}\\
	\vspace{-12pt}
	\caption{Weakly-supervised learning of the FCSS descriptor
		using correspondence consistency between object locations.}\label{img:4}\vspace{-10pt}
\end{figure}

\paragraph{Correspondence Consistency Check}
Intuitively, the correspondence relation from a source image to a target image should be consistent with that from the target image to the source image.
After forward-propagation with the training image pairs to obtain $\mathcal{F}(I;\mathbf{W})$ and $\mathcal{F}(I';\mathbf{W})$, the best match $i^*$ for each pixel $i$ is computed by comparing feature descriptors from the two images through k-nearest neighbor (k-NN) search \cite{Garcia10}:
\begin{equation}\label{equ:k-nn}
i^* = \mathop{\mathrm{argmin}}\nolimits_{i'} \| \mathcal{F} (I_{i};\mathbf{W}) - \mathcal{F} (I'_{i'};\mathbf{W}) \|^2.
\end{equation}

After running k-NN twice for the source and target images
respectively, we check the correspondence consistency and identify
the pixel pairs with valid matches as positive samples. Invalid matches are also used to generate negative samples. We randomly select the positive and negative samples during training.  Since the negative samples ensue from erroneous local minima in the energy cost, they provide the effects of hard negative mining during training \cite{Simo-Serra15b}. 
The feature learning begins by initializing the shifting
transform with randomly selected sampling patterns. We found that
even initial descriptors generated from the random patterns provide
enough positive and negative samples to be used for weakly-supervised feature learning. 
A similar observation was also reported in \cite{Kim15}.

To boost this feature learning, we limit the correspondence
candidate regions according to object location priors
such as an object bounding box containing the
target object to be matched, which are provided in most benchmarks
\cite{Fei-Fei06,Everingham10,Chen14}. Similar to \cite{Zhou15,Zhou16,Ham16}, it is assumed that true matches exist only within the object region as shown in
\figref{img:4}. Utilizing this prior mitigates the side effects that
may occur due to background clutter when directly running the k-NN, and also expedites the feature learning
process.

\section{Experimental Results and Discussion}\label{sec:4}
\subsection{Experimental Settings}\label{sec:41}
For our experiments, we implemented the FCSS descriptor using the
VLFeat MatConvNet toolbox \cite{MatConv}. For convolutional
similarity networks in the CSS layers, we used the ImageNet pretrained VGG-Net
\cite{Simonyan15} from the bottom conv1 to the conv3-4 layer, with their network parameters as
initial values. Three CSS layers are located after
conv2-2, conv3-2, and conv3-4, thus $K=3$. Considering the trade-off
between efficiency and robustness, the number of sampling patterns
is set to $64$, thus the total dimension of the descriptor is $L=192$. 
Before each CSS layer, convolutional activations are normalized to have a $L_2$ norm \cite{Song16}. To learn the network, we employed the Caltech-101 dataset \cite{Fei-Fei06} excluding testing image pairs used in experiments.  
The number of trainig samples $N$ is $1024$. $C$ is set to $0.2$. The learned parameters are used for all the experiments. Our code with pretrained parameters will be made publicly available.
\begin{table}[t]
	\centering
	\begin{tabular}{ >{\raggedright}m{0.28\linewidth}
			>{\centering}m{0.10\linewidth} >{\centering}m{0.10\linewidth}
			>{\centering}m{0.10\linewidth} >{\centering}m{0.10\linewidth}}
		\hlinewd{1.0pt}
		Methods &FD3D. &JODS &PASC. &Avg.\tabularnewline
		\hline
		\hline
		SIFT \cite{Liu11} &0.632 &0.509 &0.360 &0.500 \tabularnewline
		DAISY \cite{Tola10} &0.636 &0.373 &0.338 &0.449 \tabularnewline
		LSS \cite{Schechtman07} &0.644 &0.349 &0.359 &0.451 \tabularnewline
		DASC \cite{Kim15} &0.668 &0.454 &0.261 &0.461 \tabularnewline
		\hline
		DeepD. \cite{Simo-Serra15b} &0.684 &0.315 &0.278 &0.426 \tabularnewline
		DeepC. \cite{Zagoruyko15} &0.753 &0.405 &0.335 &0.498 \tabularnewline
		MatchN. \cite{Han15} &0.561 &0.380 &0.270 &0.404 \tabularnewline
		LIFT \cite{Yi16} &0.730 &0.318 &0.306 &0.451 \tabularnewline
		\hline
		VGG \cite{Simonyan15} &0.756 &0.490 &0.360 &0.535 \tabularnewline
		VGG w/S-CSS$^\dag$ &0.762 &0.521 &0.371 &0.551 \tabularnewline
		VGG w/S-CSS &0.775 &0.552 &0.391 &0.573 \tabularnewline
		VGG w/M-CSS &0.806 &0.573 &0.451 &0.610 \tabularnewline
		FCSS &\textbf{0.830} &\textbf{0.656} &\textbf{0.494} &\textbf{0.660} \tabularnewline
		\hlinewd{1.0pt}
	\end{tabular}\vspace{+3pt}
	\caption{Matching accuracy for various feature descriptors with fixed SF optimization on the Taniai benchmark \cite{Taniai16}. VGG w/S-CSS$^\dag$ denotes results with randomly selected sampling patterns.}\label{tab:1}\vspace{-1pt}
\end{table}
\begin{table}[t]
	\centering
	\begin{tabular}{ >{\raggedright}m{0.33\linewidth}
			>{\centering}m{0.10\linewidth} >{\centering}m{0.10\linewidth}
			>{\centering}m{0.10\linewidth} >{\centering}m{0.10\linewidth}}
		\hlinewd{1.0pt}
		Methods &FG3D. &JODS &PASC. &Avg.\tabularnewline
		\hline
		\hline
		DFF \cite{Yang14} &0.495 &0.304 &0.224 &0.341 \tabularnewline
		DSP \cite{Kim13} &0.487 &0.465 &0.382 &0.445\tabularnewline
		SIFT Flow \cite{Liu11} &0.632 &0.509 &0.360 &0.500 \tabularnewline
		Zhou \emph{et al.} \cite{Zhou16} &0.721 &0.514 &0.436 &0.556\tabularnewline
		Taniai \emph{et al.} \cite{Taniai16} &0.830 &0.595 &0.483 &0.636\tabularnewline
		Proposal Flow \cite{Ham16} &0.786 &0.653 &0.531 &0.657 \tabularnewline
		\hline
		FCSS w/DSP \cite{Kim13} &0.527 &0.580 &0.439 &0.515 \tabularnewline
		FCSS w/SF \cite{Liu11} &0.830 &\textbf{0.656} &0.494 &0.660 \tabularnewline
		FCSS w/PF \cite{Ham16} &\textbf{0.839} &0.635 &\textbf{0.582} &\textbf{0.685} \tabularnewline
		\hlinewd{1.0pt}
	\end{tabular}\vspace{+3pt}
	\caption{Matching accuracy compared to state-of-the-art correspondence techniques on the Taniai benchmark \cite{Taniai16}.}\label{tab:2}\vspace{-10pt}
\end{table}
\begin{figure}
	\centering
	\renewcommand{\thesubfigure}{}
	\subfigure[(a) FG3DCar]
	{\includegraphics[width=0.5\linewidth]{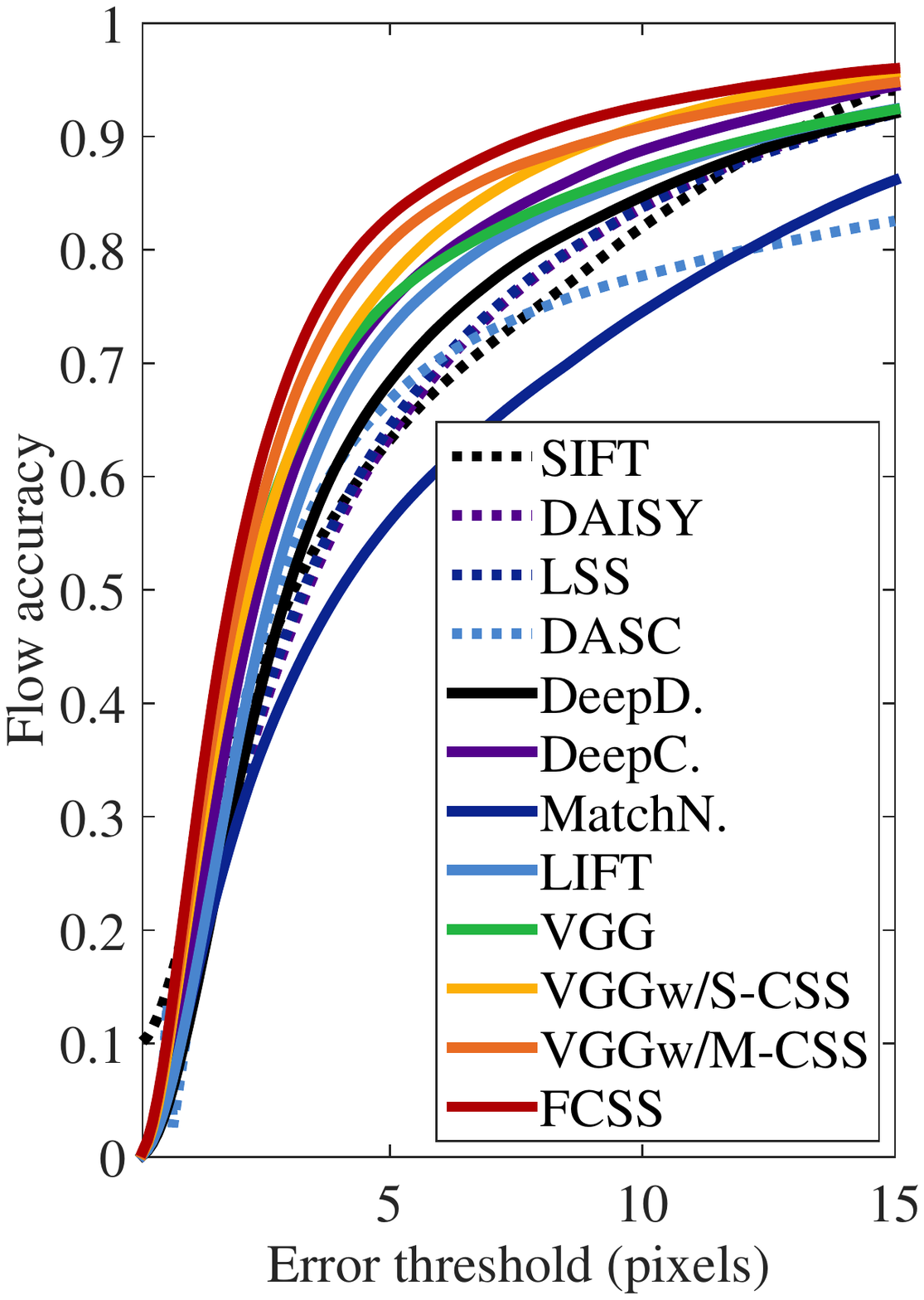}}\hfill
	\subfigure[(b) JODS]    
	{\includegraphics[width=0.5\linewidth]{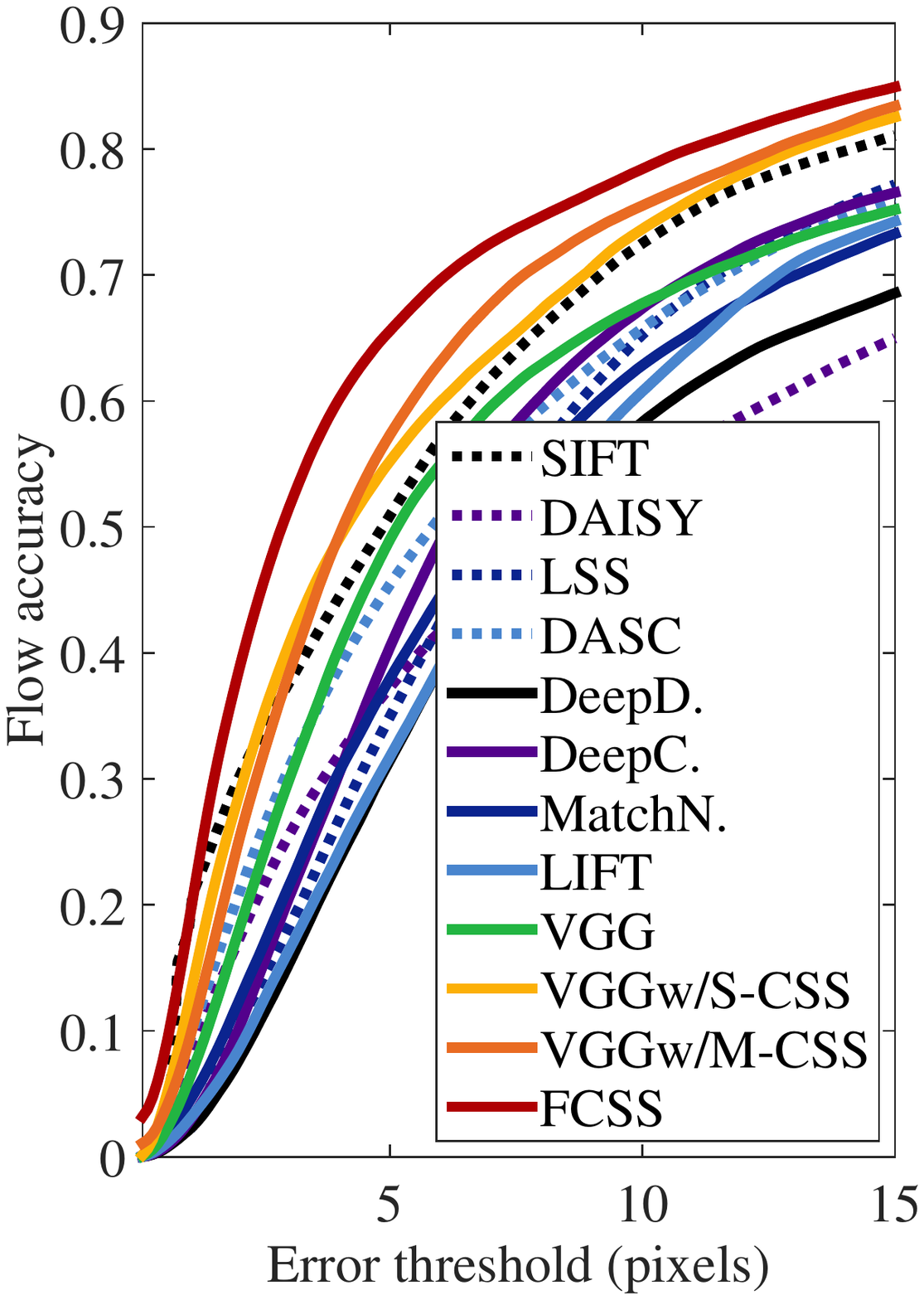}}\hfill
	\vspace{-6pt}
	\subfigure[(c) PASCAL]  
	{\includegraphics[width=0.5\linewidth]{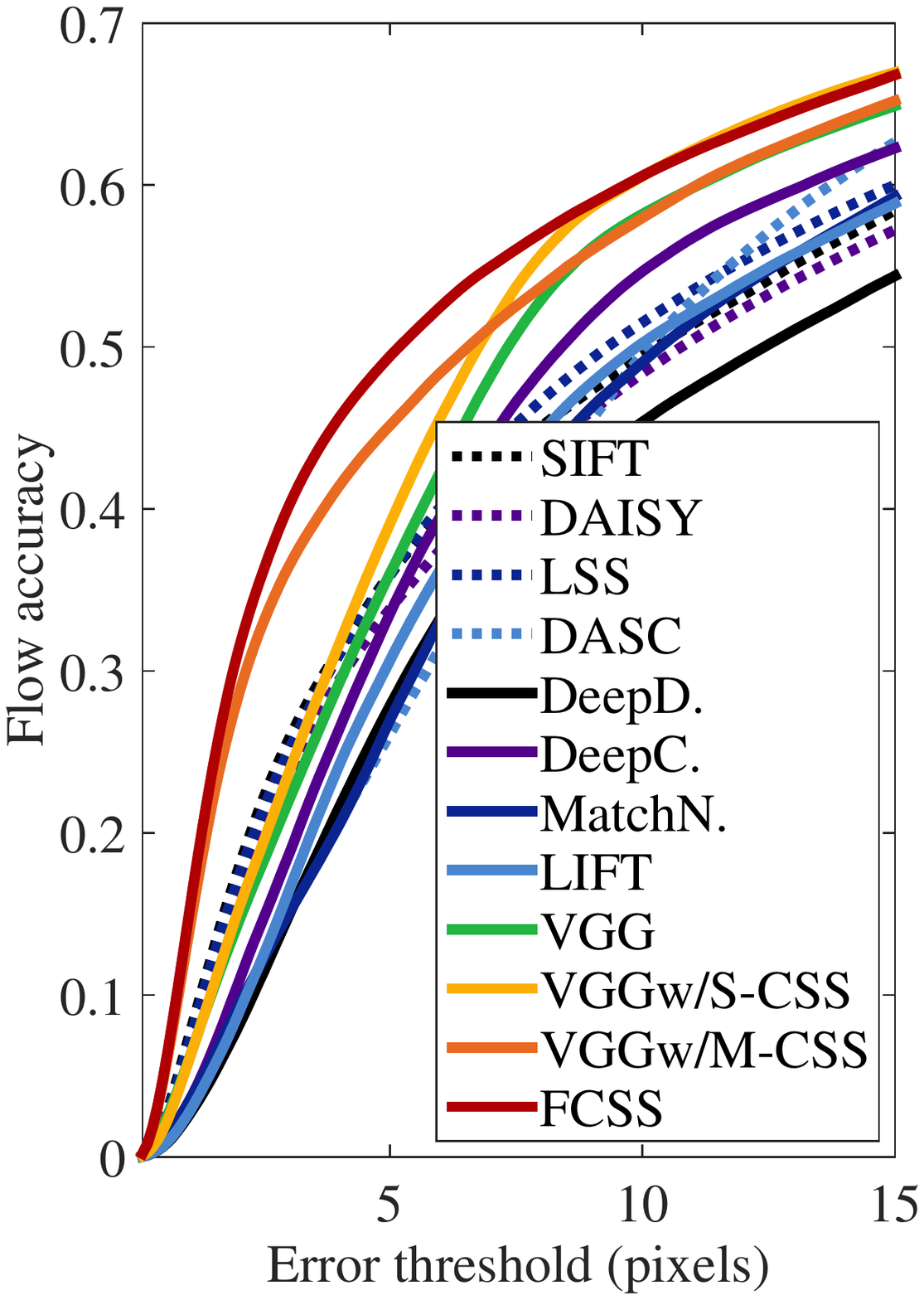}}\hfill
	\subfigure[(d) Average] 
	{\includegraphics[width=0.5\linewidth]{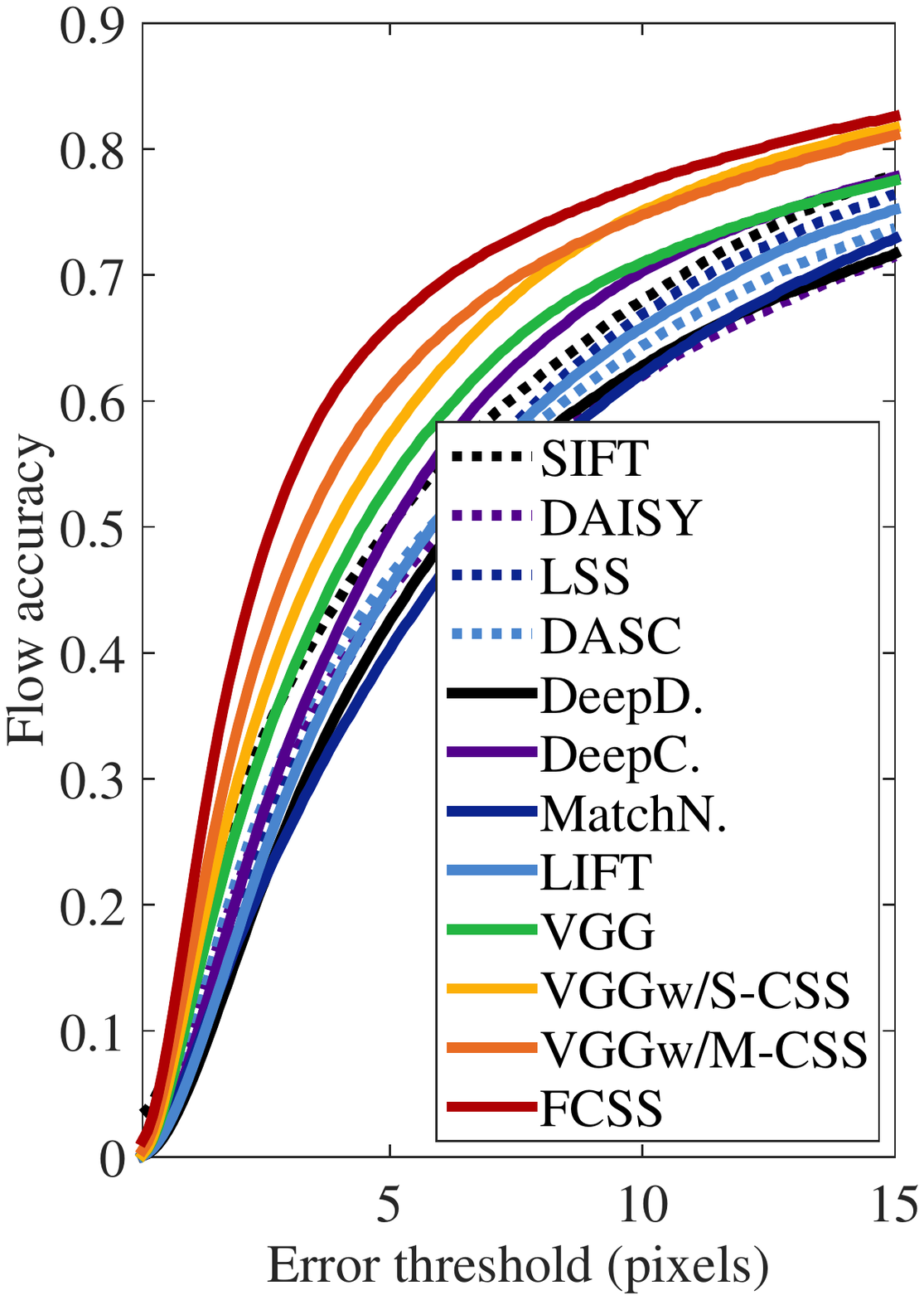}}\hfill
	\vspace{-1pt}
	\caption{Average flow accuracy with respect to endpoint error threshold on the Taniai benchmark \cite{Taniai16}.
	}\label{img:5}\vspace{-10pt}
\end{figure}

In the following, we comprehensively evaluated our descriptor through comparisons to state-of-the-art handcrafted descriptors, including SIFT \cite{Lowe04}, DAISY \cite{Tola10}, HOG \cite{Dalal05}, LSS
\cite{Schechtman07}, and DASC \cite{Kim15}, as well as recent
CNNs-based feature descriptors, including MatchNet (MatchN.) \cite{Han15}, Deep Descriptor (DeepD.)
\cite{Simo-Serra15b}, Deep Compare (DeepC.) \cite{Zagoruyko15}, UCN \cite{Choy16}, and LIFT \cite{Yi16}\footnote{Since MatchN.
	\cite{Han15}, DeepC. \cite{Zagoruyko15}, DeepD.
	\cite{Simo-Serra15b}, and LIFT \cite{Yi16} were developed for sparse
	correspondence, sparse descriptors were first built by
	forward-propagating images through networks and then upsampled.}. 
The performance was measured on Taniai benchmark \cite{Taniai16}, Proposal Flow dataset \cite{Ham16},
PASCAL-VOC dataset \cite{Chen14}, and Caltech-101 benchmark
\cite{Fei-Fei06}. To additionally validate the components of the FCSS descriptor, 
we evaluated the initial VGG-Net (conv$3$-$4$) \cite{Simonyan15} (VGG), 
the VGG-Net with learned single-scale CSS layer (VGG w/S-CSS) and learned multi-scale CSS
layers (VGG w/M-CSS)\footnote{In the `VGG w/S-CSS' and `VGG w/M-CSS',
	the sampling patterns were only learned with VGG-Net layers fixed.}.
As an optimizer for estimating dense correspondence maps, we used the
hierarchical dual-layer BP of the SIFT Flow (SF) optimization
\cite{Liu11}, whose code is publicly available. Furthermore, the
performance of the FCSS descriptor when combined with other powerful
optimizers was examined using the Proposal Flow (PF) \cite{Ham16}
and the deformable spatial pyramid (DSP) \cite{Kim13}.
\begin{figure*}
	\centering
	\renewcommand{\thesubfigure}{}
	\subfigure[]
	{\includegraphics[width=0.122\linewidth]{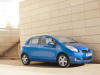}}\hfill
	\subfigure[]
	{\includegraphics[width=0.122\linewidth]{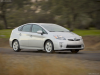}}\hfill
	\subfigure[]
	{\includegraphics[width=0.122\linewidth]{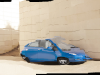}}\hfill
	\subfigure[]
	{\includegraphics[width=0.122\linewidth]{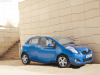}}\hfill
	\subfigure[]
	{\includegraphics[width=0.122\linewidth]{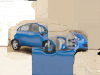}}\hfill
	\subfigure[]
	{\includegraphics[width=0.122\linewidth]{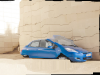}}\hfill
	\subfigure[]
	{\includegraphics[width=0.122\linewidth]{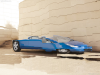}}\hfill
	\subfigure[]
	{\includegraphics[width=0.122\linewidth]{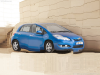}}\hfill
	\vspace{-21.5pt}
	\subfigure[(a)]
	{\includegraphics[width=0.122\linewidth]{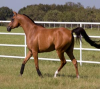}}\hfill
	\subfigure[(b)]
	{\includegraphics[width=0.122\linewidth]{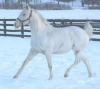}}\hfill
	\subfigure[(c)]
	{\includegraphics[width=0.122\linewidth]{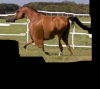}}\hfill
	\subfigure[(d)]
	{\includegraphics[width=0.122\linewidth]{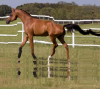}}\hfill
	\subfigure[(e)]
	{\includegraphics[width=0.122\linewidth]{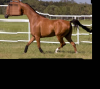}}\hfill
	\subfigure[(f)]
	{\includegraphics[width=0.122\linewidth]{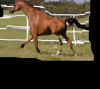}}\hfill
	\subfigure[(g)]
	{\includegraphics[width=0.122\linewidth]{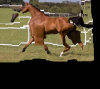}}\hfill
	\subfigure[(h)]
	{\includegraphics[width=0.122\linewidth]{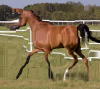}}\hfill
	\vspace{-3pt}
	\caption{Qualitative results on the Taniai benchmark
		\cite{Taniai16}: (a) source image, (b) target image, (c) SIFT \cite{Lowe04}, (d) DASC \cite{Kim15},
		(e) DeepD. \cite{Simo-Serra15b}, (f) MatchN. \cite{Han15}, (g) VGG \cite{Simonyan15}, and (h) FCSS.
		The source images were warped to the target images using correspondences.}\label{img:6}\vspace{-10pt}
\end{figure*}
\begin{figure*}[t]
	\centering
	\renewcommand{\thesubfigure}{}
	\subfigure[]
	{\includegraphics[width=0.122\linewidth]{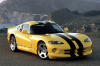}}\hfill
	\subfigure[]
	{\includegraphics[width=0.122\linewidth]{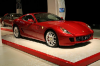}}\hfill
	\subfigure[]
	{\includegraphics[width=0.122\linewidth]{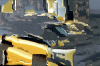}}\hfill
	\subfigure[]
	{\includegraphics[width=0.122\linewidth]{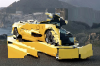}}\hfill
	\subfigure[]
	{\includegraphics[width=0.122\linewidth]{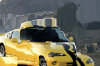}}\hfill
	\subfigure[]
	{\includegraphics[width=0.122\linewidth]{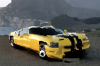}}\hfill
	\subfigure[]
	{\includegraphics[width=0.122\linewidth]{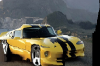}}\hfill
	\subfigure[]
	{\includegraphics[width=0.122\linewidth]{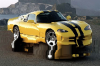}}\hfill
	\vspace{-21.5pt}
	\subfigure[(a)]
	{\includegraphics[width=0.122\linewidth]{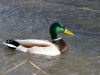}}\hfill
	\subfigure[(b)]
	{\includegraphics[width=0.122\linewidth]{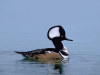}}\hfill
	\subfigure[(c)]
	{\includegraphics[width=0.122\linewidth]{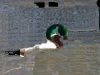}}\hfill
	\subfigure[(d)]
	{\includegraphics[width=0.122\linewidth]{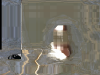}}\hfill
	\subfigure[(e)]
	{\includegraphics[width=0.122\linewidth]{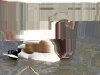}}\hfill
	\subfigure[(f)]
	{\includegraphics[width=0.122\linewidth]{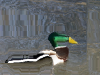}}\hfill
	\subfigure[(g)]
	{\includegraphics[width=0.122\linewidth]{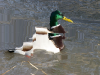}}\hfill
	\subfigure[(h)]
	{\includegraphics[width=0.122\linewidth]{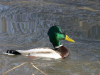}}\hfill
	\vspace{-3pt}
	\caption{Qualitative results on the Proposal Flow benchmark \cite{Ham16}: (a) source image, (b) target image,
		(c) DAISY \cite{Tola10}, (d) DeepD. \cite{Simo-Serra15b}, (e) DeepC. \cite{Zagoruyko15}, (f) LIFT \cite{Yi16}, (g) VGG \cite{Simonyan15}, and (h) FCSS.
		The source images were warped to the target images using correspondences.}\label{img:7}\vspace{-10pt}
\end{figure*}
\begin{table}[t]
	\centering
	\begin{tabular}{ >{\raggedright}m{0.27\linewidth}
			>{\centering}m{0.16\linewidth}  >{\centering}m{0.14\linewidth}
			>{\centering}m{0.16\linewidth}}
		\hlinewd{1.0pt}
		\multirow{2}{*}{Methods} &\multicolumn{3}{ c }{PCK} \tabularnewline
		\cline{2-4}
		&$\alpha=0.05$ &$\alpha=0.1$ &$\alpha=0.15$\tabularnewline
		\hline
		\hline
		SIFT \cite{Liu11} &0.247 &0.380 &0.504 \tabularnewline
		DAISY \cite{Tola10} &0.324 &0.456 &0.555 \tabularnewline
		LSS \cite{Schechtman07} &0.347 &0.504 &0.626 \tabularnewline
		DASC \cite{Kim15} &0.255 &0.411 &0.564 \tabularnewline
		\hline
		DeepD. \cite{Simo-Serra15b} &0.187 &0.308 &0.430 \tabularnewline
		DeepC. \cite{Zagoruyko15} &0.212 &0.364 &0.518 \tabularnewline
		MatchN. \cite{Han15} &0.205 &0.338 &0.476 \tabularnewline
		LIFT \cite{Yi16} &0.197 &0.322 &0.449 \tabularnewline
		LIFT$^\dag$ \cite{Yi16} &0.224 &0.346 &0.489 \tabularnewline
		\hline
		VGG \cite{Simonyan15} &0.224 &0.388 &0.555 \tabularnewline
		VGG w/S-CSS &0.239 &0.422 &0.595 \tabularnewline
		VGG w/M-CSS &0.344 &0.514 &0.676 \tabularnewline
		FCSS &\textbf{0.354} &\textbf{0.532} &\textbf{0.681}\tabularnewline
		\hlinewd{1.0pt}
	\end{tabular}\vspace{+3pt}
	\caption{Matching accuracy for various feature descriptors with SF optimization 
		on the Proposal Flow benchmark \cite{Ham16}. LIFT$^\dag$ denotes results of LIFT \cite{Yi16} with densely sampled windows.}\label{tab:3}\vspace{-14pt}
\end{table}
\begin{table}[t]
	\centering
	\begin{tabular}{ >{\raggedright}m{0.33\linewidth}
			>{\centering}m{0.16\linewidth}  >{\centering}m{0.14\linewidth}
			>{\centering}m{0.16\linewidth}}
		\hlinewd{1.0pt}
		\multirow{2}{*}{Methods} &\multicolumn{3}{ c }{PCK} \tabularnewline
		\cline{2-4}
		&$\alpha=0.05$ &$\alpha=0.1$ &$\alpha=0.15$\tabularnewline
		\hline
		\hline
		DSP \cite{Kim13} &0.239 &0.364 &0.493 \tabularnewline
		SIFT Flow \cite{Liu11} &0.247 &0.380 &0.504 \tabularnewline
		Zhou \emph{et al.} \cite{Zhou16} &0.197 &0.524 &0.664 \tabularnewline
		Proposal Flow \cite{Ham16} &0.284 &0.568 &0.682 \tabularnewline
		\hline
		FCSS w/DSP \cite{Kim13} &0.302 &0.475 &0.602 \tabularnewline
		FCSS w/SF \cite{Liu11} &\textbf{0.354} &0.532 &0.681 \tabularnewline
		FCSS w/PF \cite{Ham16} &0.295 &\textbf{0.584} &\textbf{0.715} \tabularnewline
		\hlinewd{1.0pt}
	\end{tabular}\vspace{+3pt}
	\caption{Matching accuracy compared to state-of-the-art correspondence techniques on the Proposal Flow benchmark \cite{Ham16}.}\label{tab:4}\vspace{-10pt}
\end{table}
\subsection{Results}\label{sec:42}
\paragraph{Taniai Benchmark \cite{Taniai16}}
We first evaluated our FCSS descriptor on the Taniai benchmark \cite{Taniai16},
which consists of 400 image pairs divided into three groups:
FG3DCar \cite{Lin14}, JODS \cite{Rubinstein13}, and PASCAL \cite{Hariharan11}.
As in \cite{Taniai16}, flow accuracy was
measured by computing the proportion of foreground pixels with an
absolute flow endpoint error that is smaller than a certain threshold
$T$, after resizing images so that its larger dimension is 100
pixels. \tabref{tab:1} summarizes the matching accuracy for various
feature descriptors with the SF optimization fixed
($T=5$ pixels). Interestingly, while both the CNN-based descriptors
\cite{Simo-Serra15b,Zagoruyko15,Han15,Yi16} and the handcrafted
descriptors \cite{Lowe04,Schechtman07,Tola10,Kim15} tend to show
similar performance, our method outperforms both of these
approaches. \figref{img:5} shows the flow accuracy with varying error
thresholds. \figref{img:6} shows qualitative results. 
More results are available in the supplementary
materials.

\tabref{tab:2} compares the matching accuracy ($T=5$ pixels) with other
correspondence techniques. Taniai \emph{et al.} \cite{Taniai16} and Proposal Flow \cite{Ham16} provide plausible flow fields,
but their methods have limitations due to their usage of handcrafted features. Thanks to its invariance to intra-class variations and precise localization ability,
our FCSS achieves the best results both quantitatively and qualitatively. \vspace{-10pt}
\begin{figure*}[t]
	\centering
	\renewcommand{\thesubfigure}{}
	\subfigure[]
	{\includegraphics[width=0.108\linewidth]{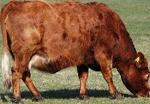}}\hfill
	\subfigure[]
	{\includegraphics[width=0.108\linewidth]{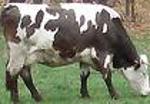}}\hfill
	\subfigure[]
	{\includegraphics[width=0.108\linewidth]{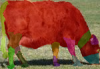}}\hfill
	\subfigure[]
	{\includegraphics[width=0.108\linewidth]{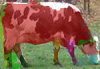}}\hfill
	\subfigure[]
	{\includegraphics[width=0.108\linewidth]{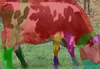}}\hfill
	\subfigure[]
	{\includegraphics[width=0.108\linewidth]{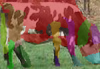}}\hfill
	\subfigure[]
	{\includegraphics[width=0.108\linewidth]{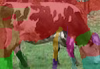}}\hfill
	\subfigure[]
	{\includegraphics[width=0.108\linewidth]{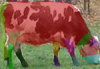}}\hfill
	\subfigure[]
	{\includegraphics[width=0.108\linewidth]{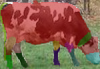}}\hfill    
	\vspace{-21.5pt}
	\subfigure[(a)]
	{\includegraphics[width=0.108\linewidth]{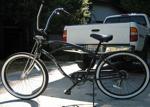}}\hfill
	\subfigure[(b)]
	{\includegraphics[width=0.108\linewidth]{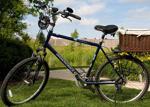}}\hfill
	\subfigure[(c)]
	{\includegraphics[width=0.108\linewidth]{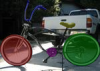}}\hfill
	\subfigure[(d)]
	{\includegraphics[width=0.108\linewidth]{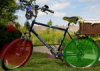}}\hfill
	\subfigure[(e)]
	{\includegraphics[width=0.108\linewidth]{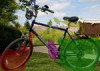}}\hfill
	\subfigure[(f)]
	{\includegraphics[width=0.108\linewidth]{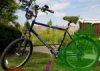}}\hfill
	\subfigure[(g)]
	{\includegraphics[width=0.108\linewidth]{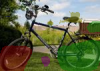}}\hfill
	\subfigure[(h)]
	{\includegraphics[width=0.108\linewidth]{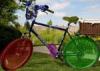}}\hfill
	\subfigure[(i)]
	{\includegraphics[width=0.108\linewidth]{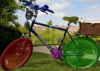}}\hfill
	\vspace{-3pt}
	\caption{Visualizations of dense flow field with color-coded part segments on the PASCAL-VOC part dataset \cite{Chen14}: (a) source image, (b) target image, (c) source mask, (d) LSS \cite{Saleem14}, (e) DeepD. \cite{Simo-Serra15b}, (f) DeepC. \cite{Zagoruyko15}, (g) LIFT \cite{Yi16}, (h) FCSS, and (i) target mask.}\label{img:8}\vspace{-10pt}
\end{figure*}
\begin{figure*}[t]
	\centering
	\renewcommand{\thesubfigure}{}
	\subfigure[(a)]
	{\includegraphics[width=0.108\linewidth]{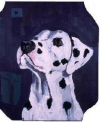}}\hfill
	\subfigure[(b)]
	{\includegraphics[width=0.108\linewidth]{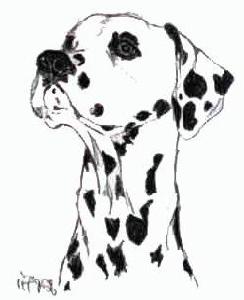}}\hfill
	\subfigure[(c)]
	{\includegraphics[width=0.108\linewidth]{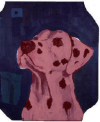}}\hfill
	\subfigure[(d)]
	{\includegraphics[width=0.108\linewidth]{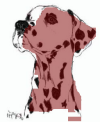}}\hfill
	\subfigure[(e)]
	{\includegraphics[width=0.108\linewidth]{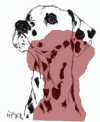}}\hfill
	\subfigure[(f)]
	{\includegraphics[width=0.108\linewidth]{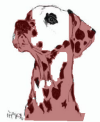}}\hfill
	\subfigure[(g)]
	{\includegraphics[width=0.108\linewidth]{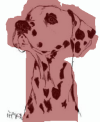}}\hfill
	\subfigure[(h)]
	{\includegraphics[width=0.108\linewidth]{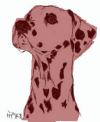}}\hfill
	\subfigure[(i)]
	{\includegraphics[width=0.108\linewidth]{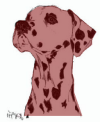}}\hfill
	\vspace{-3pt}
	\caption{Visualizations of dense flow fields with mask transfer on the Caltech-101 dataset \cite{Fei-Fei06}: (a) source image, (b) target image, (c) source mask, (d) SIFT \cite{Lowe04}, (e) DASC \cite{Kim15}, 
		(f) MatchN. \cite{Han15}, (g) LIFT \cite{Yi16}, (h) FCSS, and (i) target mask.}\label{img:9}\vspace{-10pt}
\end{figure*}

\paragraph{Proposal Flow Benchmark \cite{Ham16}}
We also evaluated our FCSS descriptor on the Proposal Flow benchmark \cite{Ham16}, 
which includes 10 object sub-classes with 10 keypoint annotations for each image.
For the evaluation metric, we used the probability of correct keypoint (PCK) between flow-warped keypoints and the ground truth \cite{Long14,Ham16}. The warped keypoints are deemed to be correctly predicted if they lie within
$\alpha \cdot \mathrm{max}(h,w)$ pixels of the ground-truth keypoints for
$\alpha \in [0,1]$, where $h$ and $w$ are the height and width of
the object bounding box, respectively.

The PCK values were measured for various feature descriptors
with SF optimization fixed in \tabref{tab:3}, and for
different correspondence techniques in \tabref{tab:4}.
\figref{img:7} shows qualitative results for dense flow estimation.
Our FCSS descriptor with SF optimization shows competitive
performance compared to recent state-of-the-art correspondence
methods. When combined with PF optimization instead, our method
significantly outperforms the existing state-of-the-art descriptors
and correspondence techniques.\vspace{-10pt}
\begin{table}[t]
	\centering
	\begin{tabular}{ >{\raggedright}m{0.33\linewidth}
			>{\centering}m{0.08\linewidth}  >{\centering}m{0.16\linewidth} >{\centering}m{0.14\linewidth}}
		\hlinewd{1.0pt}
		\multirow{2}{*}{Methods} &\multirow{2}{*}{IoU} &\multicolumn{2}{ c }{PCK} \tabularnewline
		\cline{3-4}
		& &$\alpha=0.05$ &$\alpha=0.1$\tabularnewline
		\hline
		\hline
		%Congealing \cite{Weinzaepfel13} &0.38 &0.11 &\textbf{-} \tabularnewline
		%RASL \cite{Duchenne11} &0.39 &0.16 &\textbf{-} \tabularnewline
		%CollectionFlow \cite{Kim13} &0.38 &0.12 &\textbf{-} \tabularnewline
		%DSP \cite{Kim13} &0.39 &0.17 &0.32 \tabularnewline
		%SIFTFlow \cite{Liu11} &0.39 &0.17 &0.32 \tabularnewline
		FlowWeb \cite{Kim13} &0.43 &0.26 &- \tabularnewline
		Zhou \emph{et al.} \cite{Zhou16} &- &- &0.24 \tabularnewline
		Proposal Flow \cite{Ham16} &0.41 &0.17 &0.36\tabularnewline
		UCN \cite{Choy16} &- &0.26 &0.44\tabularnewline
		%DASC \cite{Kim15} w/SF \cite{Liu11} &0.40 &0.19 &0.29\tabularnewline
		%VGG \cite{Simonyan15} w/SF \cite{Liu11} &0.42 &0.25 &0.41\tabularnewline
		\hline
		FCSS w/SF \cite{Liu11} &0.44 &0.28 &\textbf{0.47}\tabularnewline
		FCSS w/PF \cite{Ham16} &\textbf{0.46} &\textbf{0.29} &0.46\tabularnewline
		\hlinewd{1.0pt}
	\end{tabular}\vspace{+3pt}
	\caption{Matching accuracy on the PASCAL-VOC part dataset \cite{Chen14}.}\label{tab:5}\vspace{-10pt}
\end{table}

\paragraph{PASCAL-VOC Part Dataset \cite{Chen14}}
Our evaluations also include the dataset provided by \cite{Zhou15}, where the images are sampled from the PASCAL part dataset \cite{Chen14}. With human-annotated part segments, we measured part matching accuracy using the weighted intersection over union (IoU) score between transferred segments and ground truths, with weights determined by the pixel area of each part.
To evaluate alignment accuracy, we measured the PCK metric using keypoint annotations for the 12 rigid PASCAL classes \cite{Xiang14}. \tabref{tab:5} summarizes the matching accuracy compared to state-of-the-art correspondence methods. \figref{img:8} visualizes estimated dense flow with color-coded part segments. From the results, our FCSS descriptor is found to yield the highest matching accuracy.\vspace{-10pt}
\begin{table}[t]
	\centering
	\begin{tabular}{ >{\raggedright}m{0.37\linewidth}
			>{\centering}m{0.15\linewidth}  >{\centering}m{0.08\linewidth}
			>{\centering}m{0.19\linewidth}}
		\hlinewd{1.0pt}
		Methods &LT-ACC &IoU &LOC-ERR\tabularnewline
		\hline
		\hline
		%DeepFlow \cite{Weinzaepfel13} &0.74 &0.40 &0.34 \tabularnewline
		%GMK \cite{Duchenne11} &0.77 &0.42 &0.34 \tabularnewline
		DSP \cite{Kim13} &0.77 &0.47 &0.35 \tabularnewline
		SIFT Flow \cite{Liu11} &0.75 &0.48 &0.32 \tabularnewline
		Proposal Flow \cite{Ham16} &0.78 &0.50 &0.25 \tabularnewline
		%DASC \cite{Kim15} w/SF \cite{Liu11} &0.76 &0.49 &0.27 \tabularnewline
		VGG \cite{Simonyan15} w/SF \cite{Liu11} &0.78 &0.51 &0.25 \tabularnewline
		\hline
		FCSS w/SF \cite{Liu11} &0.80 &0.50 &\textbf{0.21} \tabularnewline
		FCSS w/PF \cite{Liu11} &\textbf{0.83} &\textbf{0.52} &0.22 \tabularnewline
		\hlinewd{1.0pt}
	\end{tabular}\vspace{+3pt}
	\caption{Matching accuracy on the Caltech-101 dataset \cite{Fei-Fei06}.}\label{tab:6}\vspace{-10pt}
\end{table}

\paragraph{Caltech-101 Dataset \cite{Fei-Fei06}}\label{sec:45}
Lastly, we evaluated our FCSS descriptor on the Caltech-101 dataset \cite{Fei-Fei06}.
Following the experimental protocol in \cite{Kim13}, we randomly selected 15 pairs of images for each object class, and evaluated matching accuracy with three metrics: label transfer accuracy (LT-ACC) \cite{Liu11b}, the IoU metric, and the localization error (LOC-ERR) of corresponding pixel positions. \tabref{tab:6} summarizes the matching accuracy compared to state-of-the-art correspondence methods. \figref{img:9} visualizes estimated dense flow fields with mask transfer.
For the results, our FCSS descriptor clearly outperforms the comparison techniques.

\section{Conclusion}\label{sec:5}
We presented the FCSS descriptor, which formulates local self-similarity
within a fully convolutional network. In contrast to previous LSS-based techniques, the sampling patterns and the self-similarity measure were jointly learned within the proposed network in an end-to-end and multi-scale manner. 
The network was additionally trained in a weakly-supervised manner, using correspondence consistency between object bounding boxes in the training image pairs. 
We believe FCSS can potentially benefit instance-level object detection and segmentation, thanks to its robustness to intra-class variations and precise localization ability.

%We presented the FCSS descriptor, which formulates local self-similarity within a fully convolutional network. In contrast to previous LSS-based techniques, the sampling patterns of local structure and the self-similarity measure are jointly learned within the proposed network in an end-to-end and multi-scale manner. The network is additionally trained in a semi-supervised manner, using correspondence consistency between object bounding boxes in the training image pairs. FCSS was validated on an extensive set of experiments that cover a broad range of dense semantic correspondences. We believe FCSS can potentially benefit instance-level object detection and segmentation, thanks to its robustness to intra-class variations and precise localization power.

{\small
	\bibliographystyle{ieee}
	\bibliography{egbib}
}

\end{document}